\documentclass[letterpaper, 10 pt, conference]{ieeeconf}  

\IEEEoverridecommandlockouts                              

\usepackage{graphics} 
\usepackage{epsfig} 
\usepackage{mathptmx} 
\usepackage{times} 
\usepackage{amsmath} 
\usepackage{amssymb}  
\usepackage{algorithmicx}
\usepackage{algorithm}
\usepackage{algpseudocode}
\usepackage{subcaption}
\usepackage{todonotes}
\usepackage{float}

\title{\LARGE \bf
GRP Model for Sensorimotor Learning}

\author{Tianyu Li$^1$, Bolun Dai$^2$ 
\thanks{$^1$ Tianyu Li is with the Robotics Institute, School of Computer Science, Carnegie Mellon University, 5000 Forbes Ave, Pittsburgh PA 15217, USA, {\tt\small tli3@andrew.cmu.edu}, }%
\thanks{$^2$ Bolun Dai is with the Department of Mechanical Engineering, College of Engineering, Carnegie Mellon University, 5000 Forbes Ave, Pittsburgh PA 15217, USA, {\tt\small bdai@andrew.cmu.edu}}%
}

\begin{document}

\maketitle
\thispagestyle{empty}
\pagestyle{empty}

\begin{abstract}
Learning from complex demonstrations is challenging, especially when the demonstration consists of different strategies. A popular approach is to use a deep neural network to perform imitation learning. However, the structure of that deep neural network has to be ``deep" enough to capture all possible scenarios. Besides the machine learning issue, how human learn in the sense of physiology has rarely been addressed and relevant works on spinal cord learning is rarer. In this work, we develop a novel modular learning architecture, the Generator and Responsibility Predictor (GRP) model,  which automatically learns the sub-task policies from a unsegmented controller demonstration and learns to switch between the policies. We also introduce a more physiological based neural network architecture. We implemented our GRP model and our proposed neural network to form a model the transfers the swing leg control from the brain to the spinal cord. Our result suggests that by using the GRP model the brain can successfully transfer the target swing leg control to the spinal cord and the resulting model can switch between sub-control policies automatically.

\end{abstract}

\section{INTRODUCTION}

\begin{figure*}[t]
	\centering
	\includegraphics[width=.85\textwidth]{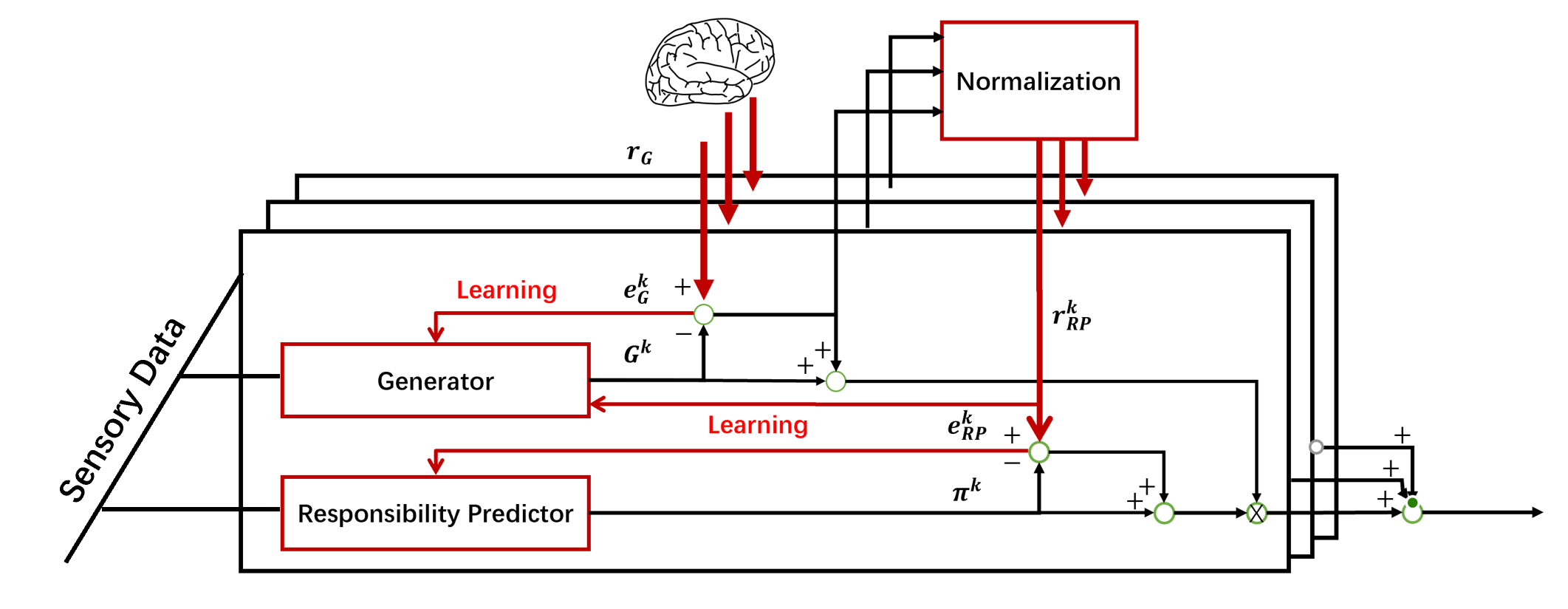}
    \caption{GRP model. The brain provides the reference control signal $r_G$ for the generators, the difference between the output of the Generator for the $k$th layer $G^k$ is sent to generate the reference signal of the RP for the $k$th layer $f^k_{RP}$. The input for our setting is the sensory data. The output of this model is equivalent to the reference control signal $r_G$.}
    \label{fig:Learning Model}
\end{figure*}


There are two distinct strategies to learn a complex task from the demonstration. The first one is to use a single deep neural network which has been widely studied in the machine learning field and implemented on computer graphics and robotics. With this approach, a single neural network can learn highly dynamic skills in simulation\cite{peng2018sfv}\cite{peng2018deepmimic}\cite{merel2017learning} and can act as the initial policy for further training that leads to deployment on real robots \cite{li2018using}\cite{zhu2018dexterous}. However, using a single monolithic policy to present a structured activity or cyclic phase structure can be challenging since a single network does not make explicit the sub-structure and encapsulates all contexts\cite{sharma2018directed}\cite{holden2017phase}. Alternatively, instead of using a monolithic controller that includes everything, a modular strategy which consists of multiple controllers and each one only is responsible for a small portion of the control. This approach has been introduced in the study of supervised learning for a mixture of demonstration data\cite{jacobs1991adaptive}\cite{jordan1992hierarchies}. These works include multiple experts' network and a classification network to split the input space into which expert are specialize according to the experts' outputs. A similar idea has been proposed which uses a directed graphical model and latent variables\cite{sharma2018directed}.

Although a variety of learning from demonstration works have been proposed, specific models that relate to human learning in physiology level is relatively rare\cite{haith2013theoretical}, and the majority of these works are focusing on the function of the brain. Gomi and Kawato\cite{gomi1993recognition} combined mixture of experts supervised learning with feedback-error learning\cite{kawato1987hierarchical}. Wolpert and Kawato, based on the idea that the brain contains multiple pairs of forward and inverse model, introduced the modular selection and identification for motor control (MOSAIC) model \cite{wolpert1998multiple}. Runbot's study was focusing on learning in the spinal cord level. Runbot investigated in the variation of gains of the neuron as gait changes using Differential Hebbian Learning\cite{wolpaw2010can}. 

Though learning occurs in the spinal cord has not been studied widely, the spinal cord plays an essential role in legged locomotion tasks\cite{courtine2009transformation}\cite{ijspeert2007swimming}. Remarkable work on mesencephalic cats and parlayed four leg mammals\cite{guertin2009mammalian}\cite{ijspeert2008central} provides direct evidence for verifying that animals can generate adaptive leg behaviors in the absence of brain planning. These observations further suggest that the spinal cord generates part of the leg control in animal legged locomotion. 

In this paper, we developed a model describing the transfer of a simple legged locomotion task: swing leg control from the brain to the spinal cord. The GRP model is an online learning model inspired by the previous work on modular architecture. We proposed a new format of the neural network which is more aligned with the biology findings in the interaction between two neurons. Then, our target swing leg controller is introduced. We tested our GRP model by learning the target swing leg control policy. The result shows that our proposed model can learn complicated controllers, and also capable of learning switch between different learned controller. Finally, we discussed our model in the physiological implications of our model and its connection to robot learning.

\section{MODELS}
The feedback error learning\cite{nakanishi2004feedback}\cite{kawato1990feedback} inspires our learning model. In the previous works, the brain uses feedback error to learn the desired command from the cerebellum. We combined the idea of feedback error learning with the modular model to proposed the GRP model. We used the GRP model to represent the transfers of a swing leg control from the brain to the spinal cord. Moreover, we introduce a new form of neural network structure which uses multiplication as the interaction between two neurons. Then we present our physical swing leg model.

\subsection{Learning Model: GRP Model}
To model the transfer of control between the brain and the spinal cord, we introduce an online learning model: Generator and Responsibility Predictor(GRP) model, which is shown in Figure \ref{fig:Learning Model}. The model is composed of multiple layers that are parallel to each other. Every layer contains a Generator and a Responsibility Predictor(RP). The generators learn from the reference control signal $r_G$ provided by the brain in our setting. The RP estimates its corresponding Generator 'responsibility' $\pi^k$ where k indicates the index of the layer. The responsibility can be interpreted as the weight of the Generator under the following constraint,
\begin{center}
    $\displaystyle\sum^m_{k=1}\pi^k = 1, \ 0\leq\pi^k\leq1$
\end{center}
RP in layer k is trained with a reference responsibility signal $r^k_{RP}$ generated by a normalization function. This normalization function takes the difference between the reference control signal and generator output $e^k_{G}$,
\begin{align*}
    e^k_G &= r_G - G^k\\
    r^k_{RP} &= \frac{e^{(-\gamma_{epi}|e^k_{G}|)}}{\sum_{i}e^{(-\gamma_{epi}|e^k_{G}|)}}\\
    \gamma_{epi} &= \beta \gamma_{epi-1} \\
    e^k_{RP} &=r^k_{RP} - \pi^k
\end{align*}
the intuition for the normalization function is to output large values when $|e^k_G|$ is small and produce small values when $|e^k_G|$ is large. $\gamma_{epi}$ is the regularization term which is a positive value and increases after every learning episode during training, the growth rate $\beta$ is a constant value larger than one. The RP error $e^k_{PR}$ is then sent backward for training the RP. 

The total output $G_{total}$ of the online learning model is,
\begin{align*}
    G_{total} &=  \sum^m_{k=1} (G^k + (r_G - G^k))(\pi^k +(r^k_{RP}-\pi^k))\\
              &= r_G\sum^m_{k=1}r^k_{RP}= r_G
\end{align*}
this guarantees that while transferring control from the brain to the spinal cord, the human is still performing locomotion in the targeting gaits. The Generator and PR are both neural network structured which will be introduced in the following section. Their inputs in our setting are sensory data, and we will discuss them in Section \ref{experiment}.

\subsection{Neural Network Structure}
\begin{figure}[]
	\centering
	\includegraphics[width=.45\textwidth]{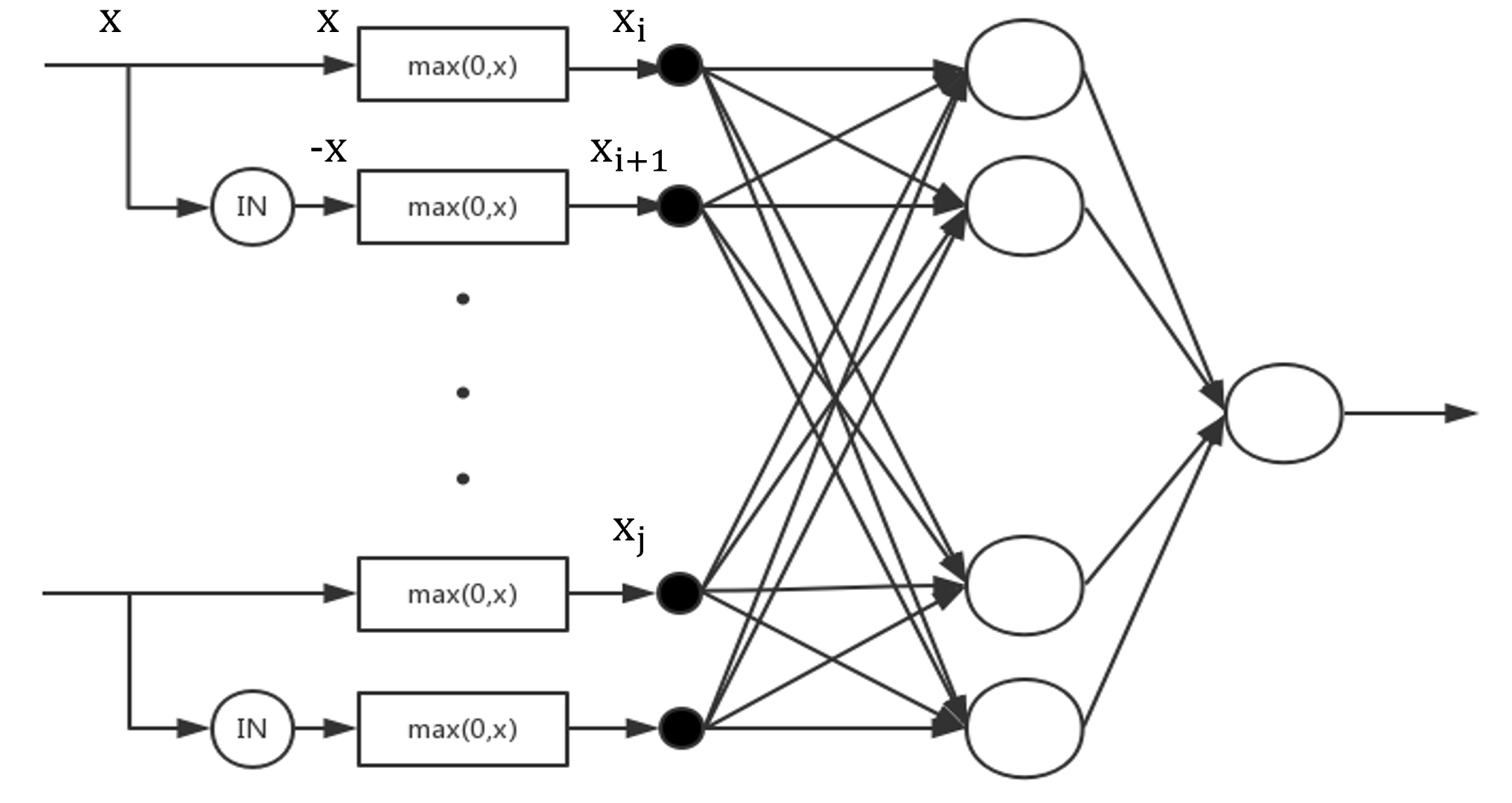}
    \caption{Neural network structure. The interaction between two neurons is represented as a multiplication. For example, the influence of neuron $j$ on $i$ is the output of neuron $i$ multiplied by $e^{w_{ij}x_{j}}$, here $w$ is the weight matrix of the neural network. When $x_j=0$ neuron $j$ will not affect neuron $i$.}
    \label{fig:neural network}
\end{figure}

Instead of using a classic neural network structure, we introduce an alternative neural network structure which enhances the biological plausibility. While class neural network models interaction between two neurons as sum operation, the typical interaction between neurons known as Presynaptic Inhibition can be better represented by multiplication. It is reasonable to assume that when one of the neurons does not sense anything, this neuron will not affect other neurons. Thus, we model the network with the interaction between two networks as the multiplication of the exponential of the input value. The network structure is shown in Figure \ref{fig:neural network}. The input data x first be sent to two tunnels and get its inverse -x in one of the channel. These two signals then pass a threshold function. This process intends to model the positive and negative part of sensory data are sensed separately by two neurons, thus provide two inputs $x_i$ and $x_{i+1}$ (one of them is a positive value, while the other must be 0) for neuron network's main part. 

The output of the Generator $G^{k}$ is
\begin{center}
    $\displaystyle G^{k} = \sum^n_{i=1}W^k_{ii}x_{i}( \prod^n_{j=1,j \neq i} e^{W^k_{ij}x_{j}})$
\end{center}
where $W^k_{ij}$ is the weight of the $j$th  input that corresponds to the $i$th input of the $k$th Generator. For RP's output $\pi^k$, we add a Sigmoid activation function to ensure the predicted responsibility value is between [0, 1].
\begin{align*}
    b^{k} &= \sum^n_{i=1}R^k_{ii}x_{i}( \prod^n_{j=1,j \neq i} e^{R^k_{ij}x_{j}})\\
    \pi^{k} &= \frac{1}{1+e^{-wb^{k}}}
\end{align*}
here, $R^k$ is the weight of the $j$th input to the $i$th input in the $k$th layer of RP.


\subsection{Brain Control Transfer}
In the proposed GRP model, the target control $r_G$ is provided by the brain. Refer to Section \ref{target control} for a detailed description of the target control. The error between the reference control and the Generator $e^k_G$ is sent backward to the Generator for learning. Meanwhile, the RP error $e^k_{RP}$ is sent back to RP for the same purpose. The weights of networks are learned via a gradient-based method using the appropriate loss function. $J^k_G$ and $J^k_{RP}$ are the loss functions for the Generator and the RP respectively,
\begin{align*}
    J^k_G &= \frac{1}{2}((e^k_G)^2 + \lambda\sum_{ij}(W^k_{ij})^2)\\
    J^k_{RP} &= \frac{1}{2}((e^k_{RP})^2 + \lambda\sum_{ij}(R^k_{ij})^2)
\end{align*}
where $\sum_{ij}(W^k_{ij})^2)$ and $\sum_{ij}(R^k_{ij})^2)$ are regularization terms, and $\lambda$ is a constant. The gradient for each weight can be computed analytically,
\begin{align*}
-\frac{\partial J^k_G}{\partial W^k_{ii}} &=  e^k_{G}  G^k  / W^k_{ii} - \lambda  W^k_{ii}\\
-\frac{\partial J^k_G}{\partial W^k_{ij}} &=  e^k_{G}  G^k  x_j - \lambda  W^k_{ij}\\
-\frac{\partial J^k_{RP}}{\partial R^k_{ii}} &= e^k_{RP}  w\pi^k(1-\pi^k) b^k / R^k_{ii} - \lambda  R^k_{ii}\\
-\frac{\partial J^k_{RP}}{\partial R^k_{ij}} &= e^k_{RP}  w\pi^k(1-\pi^k) b^k x_j - \lambda  R^k_{ij}
\end{align*}
here, $x_i$ and $x_j$ are the input data for neuron $i$ and $j$.

During learning we use the reference responsibility signals to regulate our learning rate.
\begin{center}
    $\displaystyle\mu^k=r^k_{RP}\mu$
\end{center}
here, $\mu$ is the constant learning rate, $\mu^k$ is the learning rate for the $k$th layer of the Generator at the current time step. This ensures the Generator would not learn when it takes no responsibility.

\subsection{Physical Model: Swing Leg Model}
\begin{figure}
    \centering
    \begin{tikzpicture}
\path (0, 0) coordinate (origin);
\def\radius{0.05cm}
\def\linklen{1.45cm}

\draw (origin) circle (\radius);

\draw[very thick] (origin) ++(-45:\radius) -- ++(-45:\linklen) coordinate (link1);
\path (link1) ++ (-45:\radius) coordinate (joint1);
\draw (joint1) circle (\radius);
\draw[very thick] (joint1) ++(-75:\radius) -- ++(-75:\linklen) coordinate (link2);
\path (link2) ++ (-75:\radius) coordinate (foot1) node[right, yshift=0.2cm] {$\alpha_{\rm tgt}$};
\draw[thick, -] (foot1) ++ (-0.2, 0) arc (180:120:0.2cm);
\draw (foot1) circle (\radius);

\draw[very thick] (origin) ++(-105:\radius) -- ++(-105:\linklen) coordinate (link3);
\path (link3) ++ (-105:\radius) coordinate (joint2);
\draw (joint2) circle (\radius);
\draw[very thick] (joint2) ++(-135:\radius) -- ++(-135:\linklen) coordinate (link4);
\path (link4) ++ (-125:\radius) coordinate (foot2) node[left, xshift=-0.1cm, yshift=0.2cm] {$\alpha_0$};
\draw[thick, -] (foot2) ++ (-0.2, 0) arc (180:60:0.2cm);
\draw (foot2) circle (\radius);

\draw[thin, -] (foot2) ++(-\radius, 0) -- ++(-0.5, 0);
\draw[thin, -] (foot1) ++(\radius, 0) -- ++(0.5, 0);
\path (foot2) ++(\radius, 0) coordinate (lineend);
\draw[thin, -] (foot1) ++(-\radius, 0) -- (lineend);

\draw[thin, -] (origin) ++(\radius, 0) -- ++(0.5, 0);
\draw[thin, -] (origin) ++(-\radius, 0) -- ++(-0.5, 0);

\draw[thick, dashed, ->, >=latex] (foot2) ++(60:\radius) -- ++(60:0.5) coordinate (P_mid) node[left, xshift=-0.05cm] {\tiny (i)};
\draw[thick, dashed] (P_mid) -- ++(60:0.4) coordinate (P_end);
\path (P_end) ++(60:\radius) coordinate (P) node[left, xshift=-0.1cm] {$P$};
\draw[black,fill=black] (P) circle (\radius);

\path (foot1) ++(120:\radius) ++(120:0.5) coordinate (Q_mid);
\path (Q_mid) ++(120:0.4) coordinate (Q_end);
\path (foot1) ++(120:\radius) coordinate (Q_start);
\draw[thick, dashed, ->, >=latex] (Q_end) -- (Q_mid) node[right, xshift=0.05cm, yshift=0.05cm] {\tiny (iii)};

\draw[thick, dashed, ->, >=latex] (Q_mid) -- (Q_start);
\path (Q_end) ++(120:\radius) coordinate (Q) node[above] {$Q$};
\draw[black,fill=black] (Q) circle (\radius);

\draw[thick, dashed, ->, >=latex] (P) arc (-120:-85:2cm) coordinate (arc_mid) node[below] {\tiny (ii)};
\draw (arc_mid) node[above] {$I_{clr}$};
\draw[thick, dashed] (arc_mid) arc (-85:-60:2cm) coordinate (Q);

\path (3, 0) coordinate (origin);
\def\radius{0.05cm}
\def\linklen{1.45cm}

\draw (origin) circle (\radius);

\draw[very thick] (origin) ++(-45:\radius) -- ++(-45:\linklen) coordinate (link1);
\path (link1) ++ (-45:\radius) coordinate (joint1);
\draw (joint1) circle (\radius);
\draw[very thick] (joint1) ++(-95:\radius) -- ++(-95:\linklen) coordinate (link2);
\path (link2) ++ (-95:\radius) coordinate (foot1) node[left, xshift=-0.1cm, yshift=0.2cm] {$\alpha$};
\draw[thick, -] (foot1) ++ (-0.2, 0) arc (180:110:0.2cm);
\draw (foot1) circle (\radius);

\draw[thin, -] (origin) ++(\radius, 0) -- ++(0.5, 0);
\draw[thin, -] (origin) ++(-\radius, 0) -- ++(-0.5, 0);

\path (origin) ++(-80:\radius) coordinate (beta);
\draw (foot1) ++(100:\radius) -- (beta);
\draw (origin) ++(0, -\radius) -- ++(0, -1) coordinate (beta_left) node[left] {$\beta$};
\draw (beta_left) -- ++(0, -1);
\draw[thick, -] (beta_left) arc (-90:-68:0.95cm);

\draw (origin) ++(0, \radius) -- ++(0, 0.2cm) coordinate (phi_mid) node[right, xshift=0.3cm] {$\phi_h$};
\draw (phi_mid) -- ++(0, 0.2cm);
\draw[thick, -] (phi_mid) arc (90:-48:0.2cm);

\draw[thin, -] (foot1) ++(\radius, 0) -- ++(0.5, 0);
\draw[thin, -] (foot1) ++(-\radius, 0) -- ++(-1.5, 0);

 
\draw (joint1) node[right, xshift=0.1cm] {$\phi_k/2$};
\draw (joint1) ++(-160:\radius) -- ++(-160:0.6cm);
\draw[thick] (joint1) ++(-160:0.2) arc (-160:-230:0.2);

\end{tikzpicture}
    \caption{Swing leg model. Phase 1: Contract the leg to pass the clearance length $l_{clr}$(P). Phase 2: Swing the leg to the target leg angle $\alpha_{tgt}$(Q) while holding the leg. Phase 3: Extend the leg until it reaches the ground. The right figure shows the definition of hip angle $\Phi_h$ and knee angle $\Phi_k$.}
    \label{fig:my_label}
\end{figure}
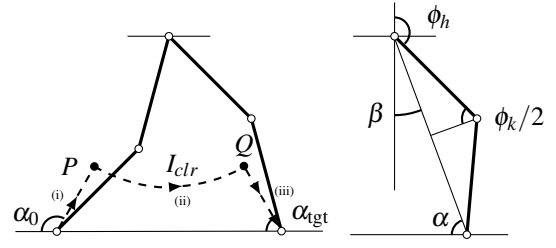

We used a classic double pendulum model as our physic model. The thigh and shank of this model are represented as rods of length $l_t$ and $l_s$ ($l_t= 50cm$, $l_s= 50cm$). Point masses $m_t$ and $m_s$ are attached to the middle point of each rod. The inertial properties are based on anthropomorphic data from a human body with the height and weight being 180cm and 80kg respectively\cite{winter2009biomechanics} ($m_t= 7.3kg$, $m_s= 4.3kg$). The hip is connected to the origin of the world frame, and the joint angles $\Phi_h$ and $\Phi_k$ are measured as shown in Figure \ref{fig:my_label}. The applied hip and knee torque $\tau_h$ and $\tau_k$ are added to hip and knee respectively. The leg angle $\alpha$ can be calculated as $\alpha = \Phi_{h} - {\Phi_k/2}$, the current leg length is calculated as $l = 2l_t\sin{\Phi_k/2}$. This model was simulated in Simulink.

\section{Target Swing Leg Controller}\label{target control}

The target swing leg controller contains three natural control tasks. Starting from the ground level at the initial configuration (leg angle $\alpha = \alpha_0$), the first task is to flex the leg to at least the clearance length $l_{clr}$. Second, the control focus shifts to advancing the swing leg to the target angle $\alpha_{tgt}$. And the final task is to extend the leg until ground contact. Although a conventional state feedback controller can execute this sequence of control, this controller takes advantage of the passive dynamics that the swing leg provides to learn the required torques.
Moreover, this controller separates the control of the hip and the knee as much as possible. As a consequence, the controller is structured around the functionally distinct hip and knee joint controllers. Overall, this swing leg control is composed of one hip control policy and three knee control policies.


\subsection{Hip Control}
The primary task of the hip controller is to move the leg to the target leg angle $\alpha_{tgt}$. The hip torque $\tau^{\alpha}_h$ is given as:
\begin{center}
    $\displaystyle\tau^{\alpha}_h = k^{\alpha}_p(\alpha_{tgt}-\alpha) - k^{\alpha}_d\Dot{\alpha}$
\end{center}
Beside angle control, the hip controller receives an additional term $\tau^{add}_h$:
\begin{center}
    $\displaystyle\tau_h =\tau^{\alpha}_h + \tau^{add}_h$
\end{center}
from the knee controller during the leg extension phase. The purpose for $\tau^{add}_h$ will be discussed in the following section.

\subsection{Knee Control}
The primary purpose of the knee controller is to regulate the leg length. As mentioned above, the controller separates knee control into three natural control tasks. Each task is assigned with an individual control policy. A detailed analysis of this controller is presented in \cite{desai2012robust}.

\subsubsection{Phase 1}
The first control task is to flexing the leg in passing a minimum clearance $l_{clr}$. The dynamics shows that while the Coriolis, centrifugal and gravitational terms always tend to extend the knee, negative hip acceleration tends to flex the knee. If the negative hip acceleration passes a threshold, no torque $\tau_k$ is required to flex the knee. Otherwise, we add an adaptive flexion control.
\begin{equation}
    \tau^i_k = \begin{cases}
        k^i\Dot{\alpha} & {\Dot{\alpha} \leq 0}\\
        0 & {\Dot{\alpha} > 0}
    \end{cases}
    \label{knee_1}
\end{equation}

\subsubsection{Phase 2}
Once the leg length has shorten pass the clearance length $l_{clr}$, the knee controller then is tasked with holding the knee, and the leg angle is only controlled by the aforementioned hip control. This task is realized when the knee flexes and modulated when the knee extends,
\begin{equation}
    \tau^{ii}_k = \begin{cases}
        -k^{ii}\Dot{\Phi}_k & {\Dot{\Phi}_k \leq 0}\\
        -k^{ii}\Dot{\Phi}_k(\alpha-\alpha_{tgt}(\Dot{\Phi}_k+\Dot{\alpha}) & {\Dot{\Phi}_k \leq 0 \And \Dot{\Phi}_k>-\Dot{\alpha}}\\
        0 & {\rm otherwise}
    \end{cases}
    \label{knee_2}
\end{equation}

\subsubsection{Phase 3}
Once the leg passes the threshold $\alpha_{thr} = \alpha_{tgt} + \Delta\alpha_{thr}$, the primary objective for the knee control switch to stopping the swing and extending leg to hit the ground. This is achieved by using two functional components. The first component generates a stopping knee-flexion torque inspired by nonlinear contact models.
\begin{equation}
    \tau^{iii}_k = \begin{cases}
        -k^{stp}(\alpha_{thr}-\alpha)(1-\frac{\Dot{\alpha}}{\Dot{\alpha}_{max}}) & {\alpha \leq \alpha_{thr} \And \Dot{\alpha}<\Dot{\alpha}_{max}}\\
        0 & {\rm otherwise}
    \end{cases}
    \label{knee_3}
\end{equation}

The stopping torque works well only if the coupling with the hip motion is canceled. Thus, we apply an a compensation torque on the hip control.
\begin{center}
    $\displaystyle\tau^{add}_h = -2\tau^{iii}_k$
\end{center}
The second functional component activates when the leg has slowed down to $\Dot{\alpha} = 0$, a knee extension torque is added to the knee to land the leg to the ground. 
\begin{center}
    $\displaystyle\tau^{iii'}_k = \tau^{iii}_k + k^{ext}(l_0-l)$
\end{center}
One thing that needs to be mention is that the last component does not necessarily needs to be activated. The swing of the leg might be terminated before the last component got activated.

\section{EXPERIMENT AND RESULTS}\label{experiment}
To exterminate the validity of our method for transferring swing leg control from the brain to the spinal cord, we used our model to learn from different trajectories generated from \cite{desai2012robust}. We tested with different numbers of Generator/Responsibility Predictor to verify that our structure is able to learn complex tasks. The inputs of the Generator and Responsibility Predictor are all sensory data, in our setting, we choose 5 variables as our system's input: $[(\alpha-\alpha_{tgt});\ \Phi_h; \ \Dot{\Phi}_{h};\ \Phi_k; \ \Dot{\Phi}_{k}]$. Since our network is sensing positive and negative values separately (see Figure \ref{fig:neural network}), the inputs to the network are 8 variables: $[(\alpha-\alpha_{tgt})^+;\ (\alpha-\alpha_{tgt})^-;\ \Phi_h;\  \Dot{\Phi}^+_{h};\ \Dot{\Phi}^-_{h};\  \Phi_k; \ \Dot{\Phi}^+_{k}; \ \Dot{\Phi}^-_{k}]$ ($\Phi_h$ and $\Phi_k$ can never be negative value, thus no $\Phi^-_{h}$ and $\Phi^-_{k}$).

\subsection{Target Control Parameters}
We used 40 different trajectories generated by a target controller for learning. The initial leg configuration is set to be the same for all trajectories: $\Phi_{h,0} = 220\ \deg$ and $\Phi_{k,0} = 175\ \deg$ ($\alpha_0 = 132.5\ \deg$). While the target leg angle is from $\alpha_{tgt} = 50\ \deg$ to $\alpha_{tgt} = 85\ \deg$ which includes the typical human landing leg angles. Initial hip velocity $\Dot{\Phi_{h,0}}$ is set to be from -4 to 0 $rads^{-1}$ while the initial knee velocity is from -7 to -1 $rads^{-1}$. The clearance leg length is $l_{clr} = 5cm$. The control gains are manually tuned, all the parameters are listed in TABLE \ref{table:para}.

\begin{table}[]
\caption{Control Parameters Values} 
\centering 
\begin{tabular}{c| c || c| c} 
\hline 
parameter & value & parameter & value \\ [0.5ex] 
\hline 
$k^{\alpha}_{p}$ & 110 & $k^{i}$ & 23 \\[0.5ex] 
$k^{\alpha}_{d}$ & 8.5 & $k^{ii}$ & 4 \\[0.5ex]
$k^{stp}$ & 250 & $\Dot{\alpha}_{max}$ & 10 \\[0.5ex]
$k^{ext}$ & 200 & $\alpha_{thr}$ & $\alpha_{tgt} + 8$ \\ [0.5ex] 
\hline 
\end{tabular}
\label{table:para} 
\end{table}

\subsection{Transfer of Control in the Neural System}
The total output of the neural network is defined as model's output when removing the reference control signal and reference responsibility, and it can be calculated as,
\begin{align*}
    \tau^{k} &= G^{k}\pi^{k}\\
    \tau_{out} &= \sum^m_{k=1}\tau^{k}
\end{align*}
here, $k$ is the index of the neural network. We modeled the transfer of hip control and knee control separately.

\subsubsection{Hip Control}
\begin{figure}
	\centering
	\includegraphics[width=.5\textwidth]{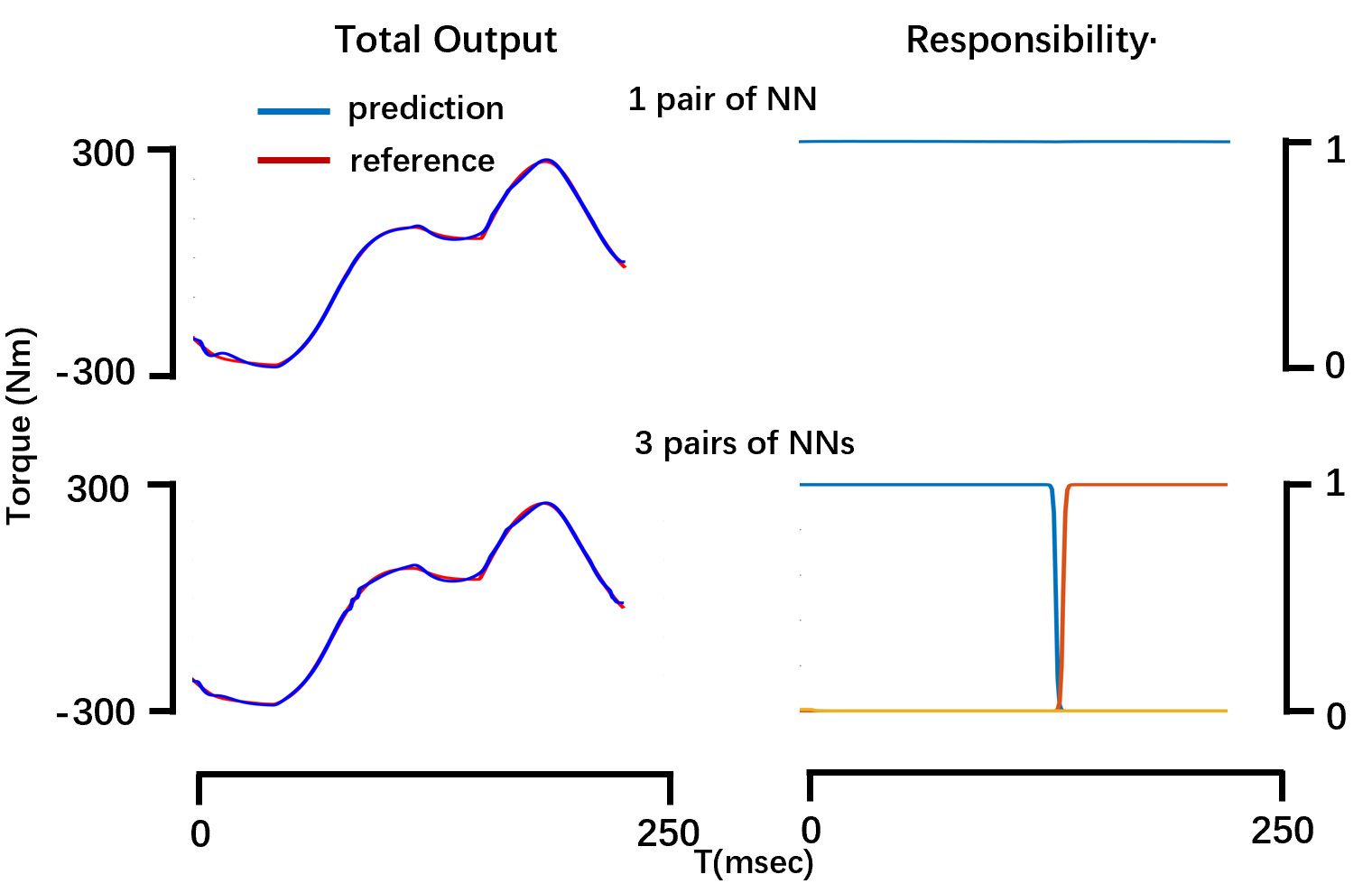}
    \caption{Result of learning the hip control using a single pair of networks (Generator and RP) and 3 pairs of networks. The left graph shows the total output. The right graph shows the predicted responsibility, each color represents a different Generator.}
    \label{fig:hip result}
\end{figure}

With used a pair of networks to train for the hip control initially. As shown in Figure \ref{fig:hip result}, using a single network, the prediction (blue) overlaps with the target control output (red), we can conclude that the system learned the target control and this indicates the control transferred to the spinal cord. The resulting RP is equal to 1 during the control that is because, in this setting, there is only one pair of networks activated, the corresponding Generator should always be in charge of the command, which means at any time the responsibility should output $100\%$. We also tested using three pairs of networks to model hip control transfer. After training, the total output fits the target well and slightly better than the previous setting. Although there are three paralleled Generator/RPs the resulting control only activated 2 of them in the test swing while one of them remained silent. We will discuss this formally in the knee control section.

\subsubsection{Knee Control}
\begin{figure}[h]
	\centering
	\includegraphics[width=.5\textwidth]{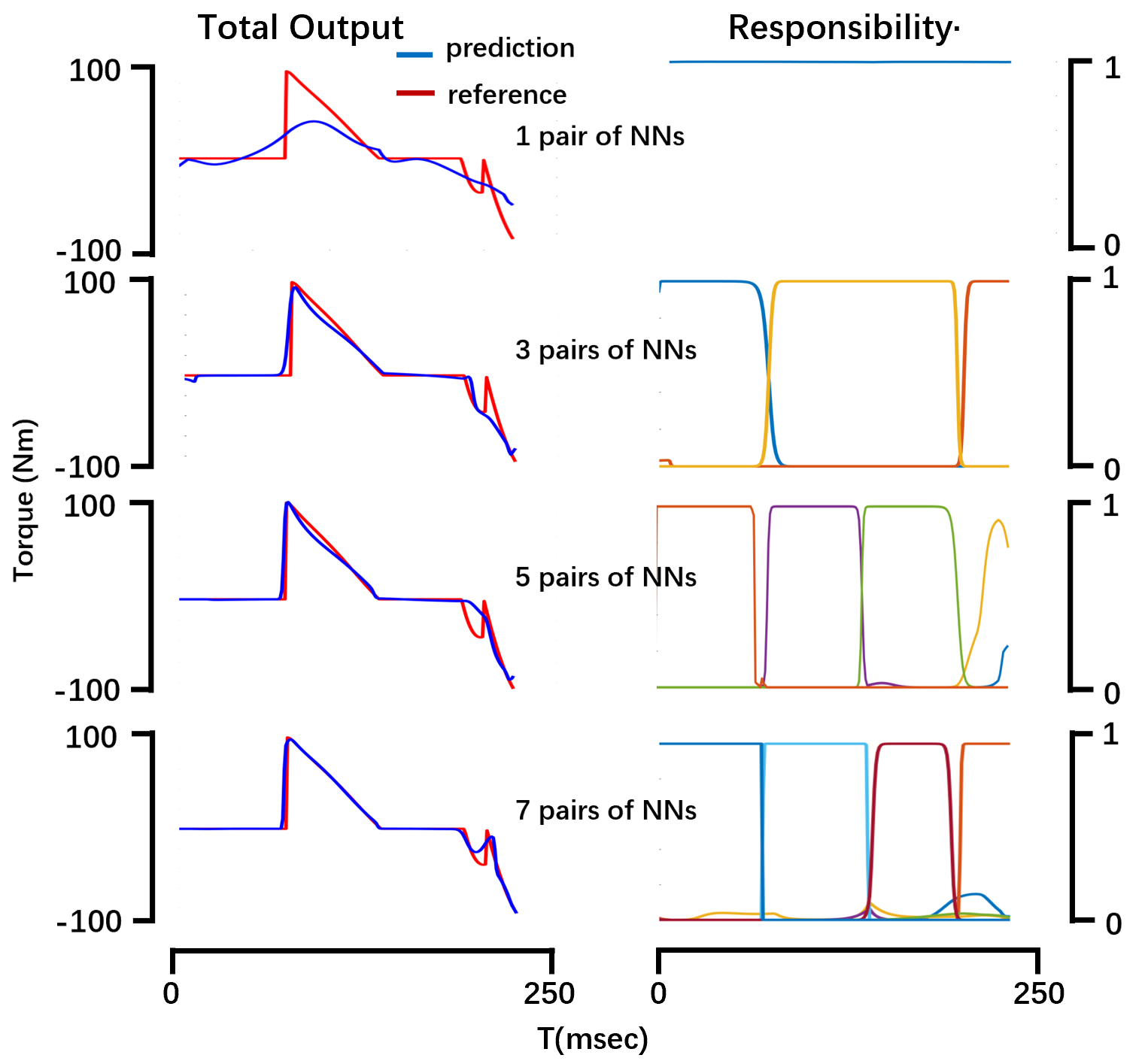}
    \caption{Result of learning the knee control using 1, 3, 5, 7 pairs of networks (Generator and RP). The left graph shows the total output. The right graph shows the predicted responsibility, each color represents a different Generator.}
    \label{fig:knee result}
\end{figure}

For the target knee control, it consists of three distinct control objectives and each objective is formed by an individual control strategy. Thus, we can conclude that the target hip control is harder to learn. We started with using a single pair of networks, from the result (Figure \ref{fig:knee result}) we can see that due to the complexity of the target control, the final result cannot match the target entirely. Then, we used multiple pairs of networks to learn the target control. As the number of networks increased, the prediction can fit the target control output better. In the three pairs of networks setting, the GRP model learned three controllers and learned to switch between them. When using five and seven pairs of networks, in Figure \ref{fig:knee result}, the GRP in both of the settings found 4 primary controllers activated during the test. According to the resulting figure, the last two settings, even though they have five and seven Generators respectively, they only enabled four of them. Moreover, these two settings learned to switch controller at the same time steps. Combining the results of hip learning and knee learning, we can conclude that the GRP model can select the number of network pairs automatically instead of activating all the layers.

\subsection{Identified Controller}
\begin{figure}[ht]
	\centering
	\includegraphics[width=.4\textwidth]{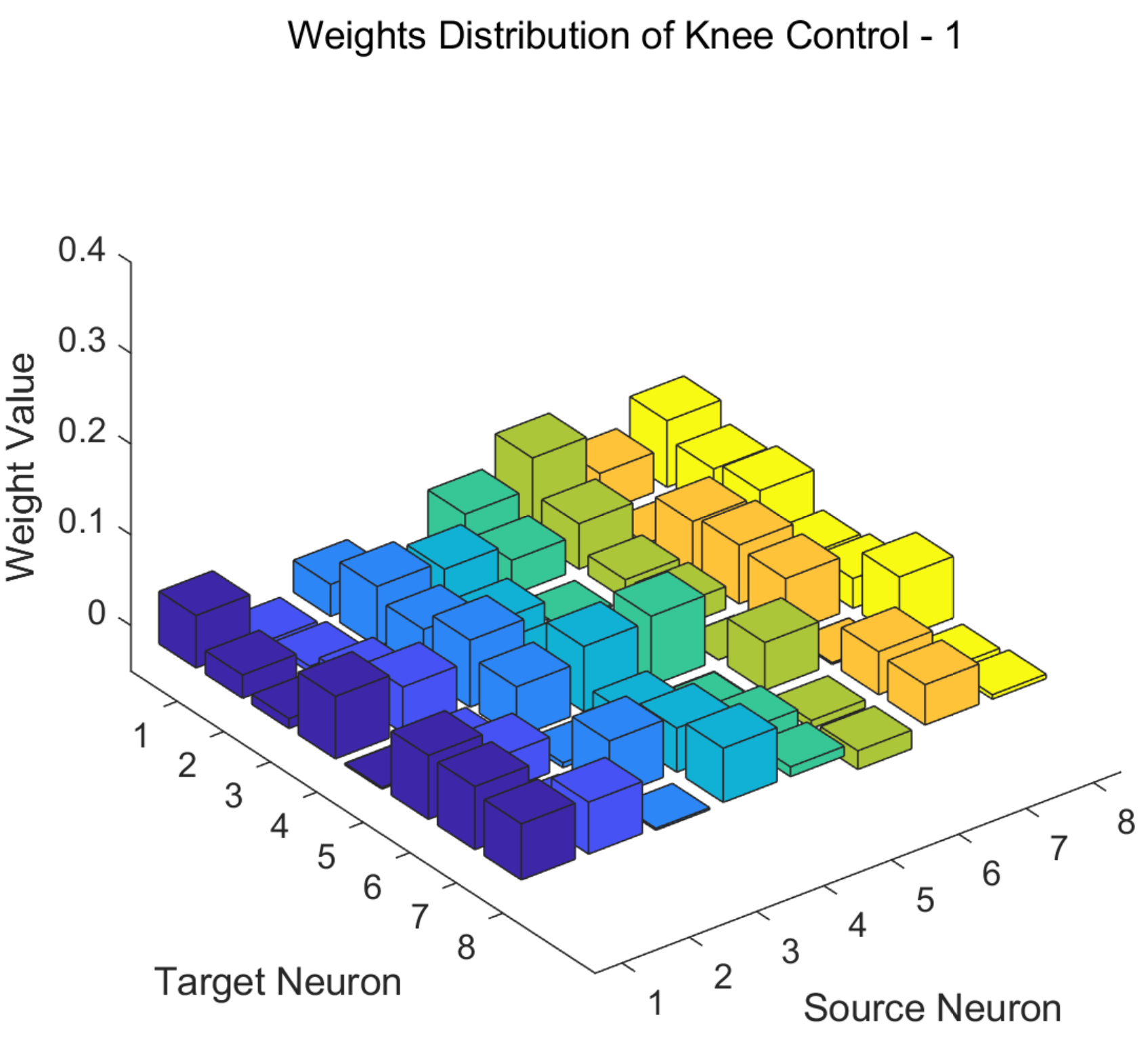}
    \caption{The indices 1-8 represents $[(\alpha-\alpha_{tgt})^+;\ (\alpha-\alpha_{tgt})^-;\ \Phi_h;\  \Dot{\Phi}^+_{h};\ \Dot{\Phi}^-_{h};\  \Phi_k; \ \Dot{\Phi}^+_{k}; \ \Dot{\Phi}^-_{k}]$ respectively. Different colors represents different source neurons. The $x$-axis represents the source neuron, the $y$-axis represents the target neuron and the $z$-axis represents the weight value. }
    \label{fig:WD}
\end{figure}

Looking into the learned weights distribution of networks, we can find a controller that explainable. Extracting the weights of the first responsible Generator of knee control in the multiple layers setting, we get weights distribution Figure \ref{fig:WD}. From the weights distribution, we can identify a 'passive' Generator since those weights are close to 0 which means given any inputs the output of this generator close to 0. 

\subsection{Resulting Spinal Cord Control Performance}
We tested the overall performance of the resulting model by removing all the reference signals. We used the resulting single pair network for hip control and 3 pairs of networks for knee control. The performance is defined by the average error and the maximum error. The error here is the absolute value of $\alpha_{tgt}-\alpha_{end}$ which is the distance between the target leg angle and the actual final leg angle $\alpha_{end}$. We sampled 20 trajectories using the same range of initial condition and target angle as the training data, with an average of $4.8 \deg$ error and a maximum error of $9.4 \deg$ (the target controller has an average error of $2.5 \deg$ and a maximum error of $6.2 \deg$).

\section{DISCUSSION}
\subsection{Transfer of Control in the Neural System}
Our result suggests a framework where the neural system could be used to transfer the control from brain to spinal cord level. However, some points need to be clearly stated.

Firstly, the model assumes the learning occurs in human which requires the neurons to be able to compute the error and propagate the error back to the reflex synapses for learning. While there is evidence showing the existence of back-propagation within neurons, there is no prove showing that every neuron has such ability. If the neurons connected between brain and the spinal cord is not able to do back-propagation our approach would be physiological implausible.

The second point is that, although we enhance our physiological implausibility by creating a norm form of neural network structure, this structure still need improvement. During sensing, neurons have threshold to filter the low sensing data. This effect can be represented by adding a bias in our current structure. The bias can also be comprised in our learning process as variables.

Moreover, although our network considers the relationship between two neurons as multiplication in order to model the Presynaptic Inhibition which is commonly found in neuron systems, our swing leg model is a simple double pendulum which can extend to a muscle-skeletal model\cite{song2013integration}\cite{desai2013muscle}. Specifically, the knee and ankle torques are generated by related Hill-type muscles, e.g. gastrocnemius. Each muscle produces a forces as a function of the muscle's current stimulation, the muscle length and the muscle velocity \cite{geyer2003positive}. By investigating the network relationship between each muscle's stimulation and positive force/length/velocity feedback from different muscles, we might be able to understand how this adaptation shapes the controller structure in the muscle level.

\subsection{Learning Framework}
Beside the biological plausibility, the structure, and its learning algorithm, although learned the target controller in our setting and can learn to switch control between sub-controls, there are a few points that are worth to be addressed which might inspire new ideas in the robot learning community.

First of all, by increasing the number of pairs of networks, the capability of the system will increase correspondingly, which indicates the ability to learn any control using demonstrations. Plus, there is no doubt that the network structure in our setting can be replaced by other forms like multi-layer perceptions (MLP). For a more complex task which composes a series of sub-tasks, our learning algorithm can distinguish different sub-tasks, this shows the potential of this method in AI reasoning field, as a comparison, current state-of-the-art techniques, using deep neural networks to fit the target control suffers from low explainability. These identified different controllers can be stored in a 'skill library'; those stored controllers can be used separately to achieving complex tasks that consist of several learned sub-tasks. Besides, each controller can be optimized independently using individual well-designed cost functions.

Besides that, our method is based on the assumption that each generator network is initialized with different weights if two sets of weights are equal, then at each step since the error between the output and the target are identical, they will have the same update, which leads to having the weights. Besides the situation as mentioned earlier, in a more general case, where two sets of weight are similar, this would lead to a slow learning process. In other words, the initialization weights are crucial to our algorithm. Since in our setting, we use a simple gradient descent method to optimize the error between prediction and target, in the MOSAIC model, a similar issue has been alleviated using the EM algorithm and hidden Markov model (HMM) based learning\cite{haruno2001mosaic}, thus, we can try to ease this issue by replacing gradient descent with other optimization methods.

\section{CONCLUSION}
We proposed a novel neural network architecture that uses multiplication as the primary relationship between neurons to imitate the Presynaptic Inhibition effect in neural systems. To use multiple simple neural network structure to learn from an unsegmented complex control policy, we used a predefined multi-phases swing leg control. We introduced a modulate model which is composed by several groups of generator and responsibility predictors. The generator predicts the current output and the responsibility predictor generates the weight of its corresponding generator. Using this model, we modeled the transfer of a swing leg control from the brain to the spinal cord level. We demonstrated that this model can learn when to switch on/off the generator and is able to automatically select the number of Generator it uses. We discussed our work in the context of both physiology and robot learning. For physiology, we can further extend this work by using more physical models like the muscle model which might ignite new physiology findings. For robot learning, we show the potential of this model in AI reasoning. We also discussed some improvements that can be implemented to our model.

\bibliographystyle{IEEEtran}

\bibliography{reference}
\end{document}